\documentclass[runningheads]{llncs}
\usepackage{amsmath,graphicx,caption,mathptmx,fixltx2e,mathtools
,amssymb}
\usepackage{subcaption}
\begin{document}
\title{Blind Deconvolution Microscopy Using Cycle Consistent CNN with Explicit PSF Layer}
%
%
\author{Sungjun Lim\inst{1}\and
Sang-Eun Lee\inst{2}\and
Sunghoe Chang\inst{3}\and
Jong Chul Ye\inst{4}
}

\authorrunning{SJ et al.}
%
\institute{KAIST Institue for Artificial Intelligence, Daejeon, Republic of Korea \and
Dept. of Physiology \& Biomedical Sciences
Seoul National University College of Medicine, Seoul, Republic of Korea \and
Dept. of Physiology \& Biomedical Sciences
Seoul National University College of Medicine, Seoul, Republic of Korea \and
Dept. of Bio/Brain Engineering, KAIST, Daejeon, Repulic of Korea
}

\maketitle              
\begin{abstract}
Deconvolution microscopy has been extensively used to improve the resolution of the widefield fluorescent microscopy. Conventional approaches, which usually  require the point spread function (PSF) measurement or blind estimation,
are however computationally expensive.  Recently, CNN based approaches have been explored as a fast and high performance alternative. In this paper, we present a novel unsupervised deep neural network  for blind deconvolution based on cycle consistency and PSF modeling layers. In contrast to the recent CNN approaches for similar problem, the explicit PSF modeling layers improve the robustness of the algorithm. Experimental results confirm the efficacy of the algorithm.
\end{abstract}
\section{Introduction}
In fluorescent microscopy,
light diffraction from a given optics degrades the resolution of  images.
To improve resolution, many optimization-based deconvolution algorithms have been developed \cite{chaudhuri2014blind,sarder2006deconvolution,mcnally1999three}. 
When the PSF measurements are not available,
You et al. \cite{you1996regularization} proposed a blind deconvolution method by solving joint minimization problem to estimate the unknown
blur kernel and  the image. Chan et al. \cite{chan1998total} proposed an improved version of blind deconvolution
using {TV} regularization.

{
Recently, convolutional neural networks (CNN)  have been extensively used to enhance peformance of an optical microscope without  hardware changes. 
 Rivenson et al. \cite{rivenson2017deep} used deep neural networks to improve optical microscopy, enhancing its spatial resolution over a large field of view and depth of field. Nehme et al. \cite{nehme2018deep} used deep convolutional neural network that can be trained on simulated data or experimental measurement to obtain super resolution images from localization microscopy. 
 Weigert et al. \cite{weigert2017isotropic} proposed CNN method which can recover isotropic resolution from anisotropic data. In addition, generative adversarial network (GAN) has attracted much attention in inverse problem by providing a way to use unlabeled data to train a deep neural network \cite{mccann2017convolutional}. Kupyn et al. \cite{kupyn2018deblurgan} presented DeblurGAN for motion deblurring using a conditional GAN and content loss. However,  this GAN approaches often generates the artificial features due to the mode collapsing, so
 a cycle-consistent adversarial network (CycleGAN) \cite{zhu2017unpaired} that imposes the one-to-one correspondency
has also made impact on image reconstruction \cite{kang2019cycle,lu2017conditional}.
 
However, these CycleGAN approaches usually require two generators with high capacity, which are often difficult to train with small
number of training data.  To address this problem, this paper proposes a novel CycleGAN architecture with an explicit PSF layer for blind
deconvolution problems. Thanks to the simple PSF layer that generates blur images, 
  we show that our proposed method is robust and efficient for the deconvolution task in spite of fully exploiting the cyclic consistency for blind deconvolution.} 

\begin{figure}[h!]
\centering\includegraphics[width=0.8\textwidth]{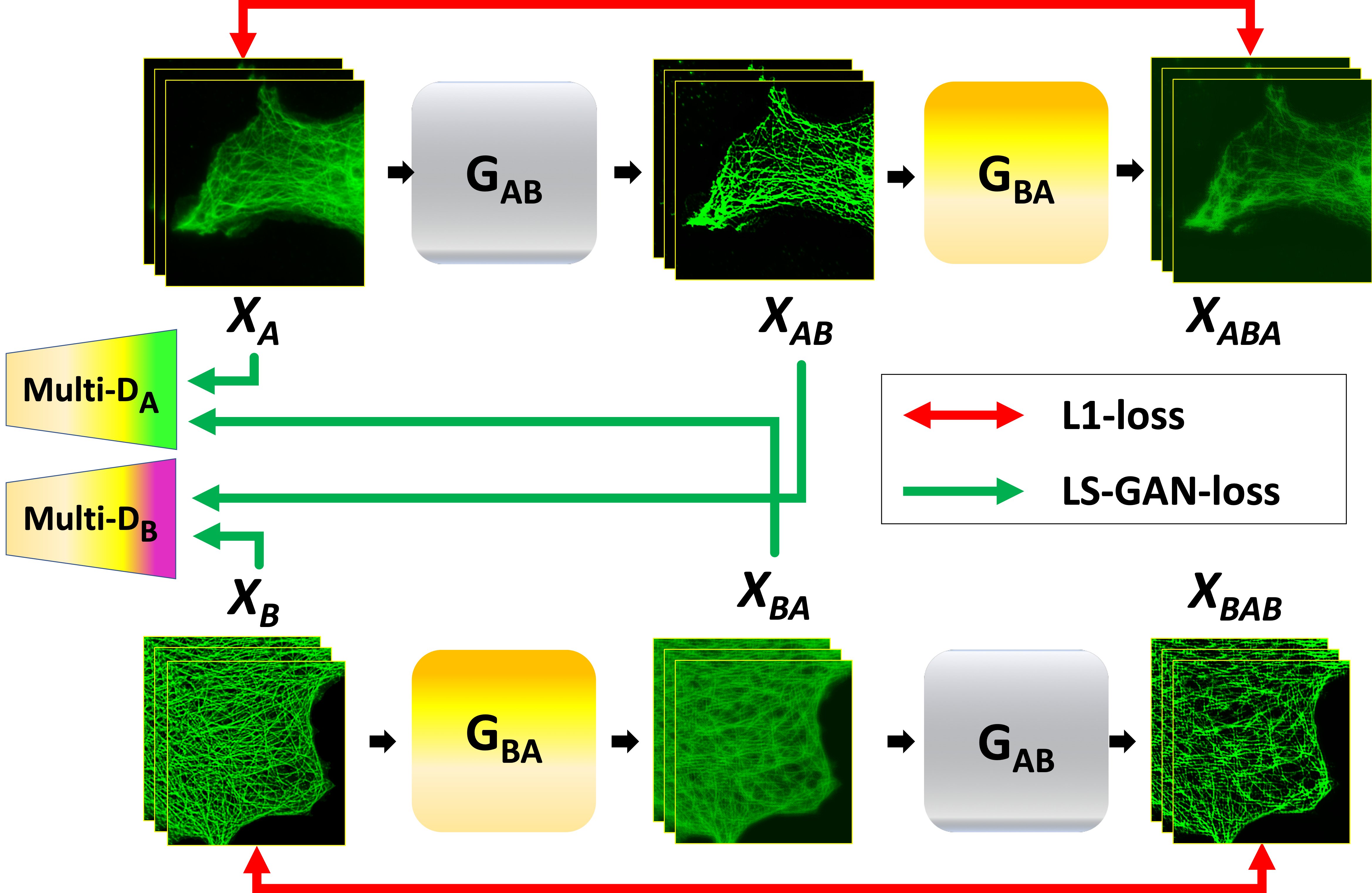}
\caption{Overall architecture of our proposed method. $G_ {AB} $ are generators that map the blur domain to the sharp image domain, and  $ G_ {BA} $  is an explicit PSF layer that needs to be estiamted.  Multi-$D_{A}$,$D_{B}$ are modules that contain independent discriminators which take cropped patches on different scale. }
\label{fig:overallCyc_framework}
\end{figure}

\section{Theory}
Fig.\ref{fig:overallCyc_framework} illustrates overall framework of the proposed method. We refer to  $A$ as the blurred
image domain and  $B$ as the blur-removed sharp image domain. The generator  $G_{AB}$  then
 maps a blurred image in  $A$ to a sharp image in  $B$, and the generator $G_{BA}$ corresponds to blur operation from
sharp image domain $B$ to a blurred measurement domain $A$. In contrast to the existing cycle-GAN architecture for blind deconvolution \cite{lu2017conditional}, we use an explicit PSF layer for
the map $G_{BA}$, in which the actual PSF values are estimated from the training data.
 
 While  the use of an explicit PSF layer can
 have a risk to reduce the generalizability of the PSF, we found that in typical microscopic setups with predetermined optics,
 the PSF is generally
 fixed so that sample-dependent PSF adaptation is not  much necessary. 
 Instead, the use of an explicit PSF layer significantly improves the stability of the algorithm.
 
%
%
%
  In addition,  the discriminator network $D_{A}$  is designed to distinguish the synthetically generated blurred image
  from  real ones. Similarly,  ${D}_{B}$ is to discriminate  generated deblurred images from  sharp image distribution.
  For the sharp image distribution, we use {\em un-matched} high resolution images. These could come from
  super-resolution microscopy or from commercially available deconvolution software.
Finally, we train both the generators and the discriminators in an alternating manner by solving the following optimization problem: 
\begin{equation}
\label{eq:1}
\underset{G_{AB},G_{BA}}{\min}   \underset{D_{A},D_{B}}{\max}L(G_{AB},G_{BA},D_{A},D_{B})
\end{equation}
in which the loss function is defined as follows:
\begin{eqnarray*}
L(G_{AB},G_{BA},D_{A},D_{B})&=&L_{GAN}(G_{AB},D_{B},A,B)+ L_{GAN}(G_{BA},D_{A},B,A)\\ 
&&+ \lambda_{1} L_{cyclic}(G_{AB},G_{BA}) + \lambda_{2}\|G_{BA}\|_{1}
\end{eqnarray*}
where $\lambda_{1}, \lambda_2$ are  hyperparameters, and $L_{GAN}$, $L_{Cyclic}$ are an adversarial loss, cyclic loss respectively. $\|G_{BA}\|_1$ is the L1-norm for the regularization of blur kernel. In following sections, we will give further explanation regarding each component of the loss function.

\subsection{Loss function}
\subsubsection{Adversarial loss}
We employed the modified GAN loss using a Least Square Loss \cite {mao2017least}. Specifically,  the min-max optimization problem for GAN training is composed of   two separate minimization problems as follows:
\begin{equation}
\begin{split}
&\underset{G_{AB}}{\min}\mathbb{E}_{x_{A} \sim P_{A}}\bigg[(D_{B}(G_{AB}(x_{A}))-1)^2 \bigg]
\end{split}
\end{equation}
\begin{equation}
\begin{split}
&\underset{D_{B}}{\min}\frac{1}{2}\mathbb{E}_{x_{B} \sim P_{B}} \bigg[(D_{B}(x_{B})-1)^2\bigg] + \frac{1}{2}\mathbb{E}_{x_{A} \sim P_{A}} \bigg[D_{B}(G_{AB}(x_{A}))^2\bigg] 
\end{split}
\end{equation}
where $ P_ {A} $ and $ P_ {B} $ denote the distribution for the domain $A$ and $B$.
By optimizing the adversarial loss, we can regulate the generators so that the generated sharp image volume is as realistic as possible; at the same time, the discriminators are optimized to distinguish the generated deconvoluted image volume from the real high resolution image. The equivalent adversarial loss was also imposed on $G_{BA}$ for deceiving generation of synthetic blurred data. 

\subsubsection{Cyclic loss}
Although mapping between $(A)$ and $(B)$ can be estimated by a well trained adversarial network, it is still vulnerable to the mode failure problem in which many input images are taken into a fixed output image. Also, because of the large capacity of a deep neural network, the network can map $(A)$ to any random permutation of the output in the domain $(B)$ that the target distribution is likely to match. In other words, the adversarial loss alone cannot guarantee a reversal of both domains. In order to resolve such issues, Zhu  {et al}. \cite{zhu2017unpaired} proposed  {cycle consistency} loss. In our case, the loss of   {cycle consistency} supports a one-to-one correspondence between the blurred image volume and the deconvoluted volume. The specific {cycle consistency} loss is defined as follows:
\begin{equation}
\begin{split}
 {L}_{cyclic}(G_{AB},G_{BA}) = & \mathbb{E}_{x_{A} \sim P_{A}}\bigg[\|G_{BA}(G_{AB}(x_{A}))-x_{A}\|_{1}\bigg] + \mathbb{E}_{x_{B} \sim P_{B}}\bigg[\|G_{AB}(G_{BA}(x_{B}))-x_{B}\|_{1}\bigg] 
\end{split}
\end{equation} 

\subsection{Multi patchGANs in CycleGAN }

As for the discriminators, we propose an improved model from the original CycleGAN using multi-PatchGANs (mPGANs) \cite {isola2017image}, where each discriminator has input patches with different sizes used. PatchGAN typically focuses on high-frequency structures by including local patches for the entire image. 
Because patches with different scales can contain different high-frequency structures, we use multiple discriminators that take the patches at different scales. Specifically, we define multi-discriminator as $\{D^{f_{i}}_{A},D^{f_{i}}_{B}\}$ where $f_{i}$ denotes the
$i^{th}$ scale patch. 
The adversarial loss with the multiscale patches is then formulated as follows:
\begin{equation}
\begin{split}
L_{GAN}(G_{AB},D_{B},A,B) = &\mathbb{E}_{X_{B} \sim P_{B}}{\bigg[ \sum_{\mathclap{i=1}}^{N}\bigg(1-D^{f_{i}}_{B}(X_{B})\bigg)^2\bigg]} +\mathbb{E}_{X_{A} \sim P_{A}}{\bigg[ \sum_{\mathclap{i=1}}^{N}\bigg(D^{f_{i}}_{B}(G_{AB}(X_{A}))\bigg)^2 \bigg]}
\end{split}
\end{equation}
where $N$-denotes the number of total scales.
$L_{GAN}(G_{BA},D_{A},B,A)$ is similarly defined.

\begin{figure}[h!]
\centering\includegraphics[width=0.8\textwidth]{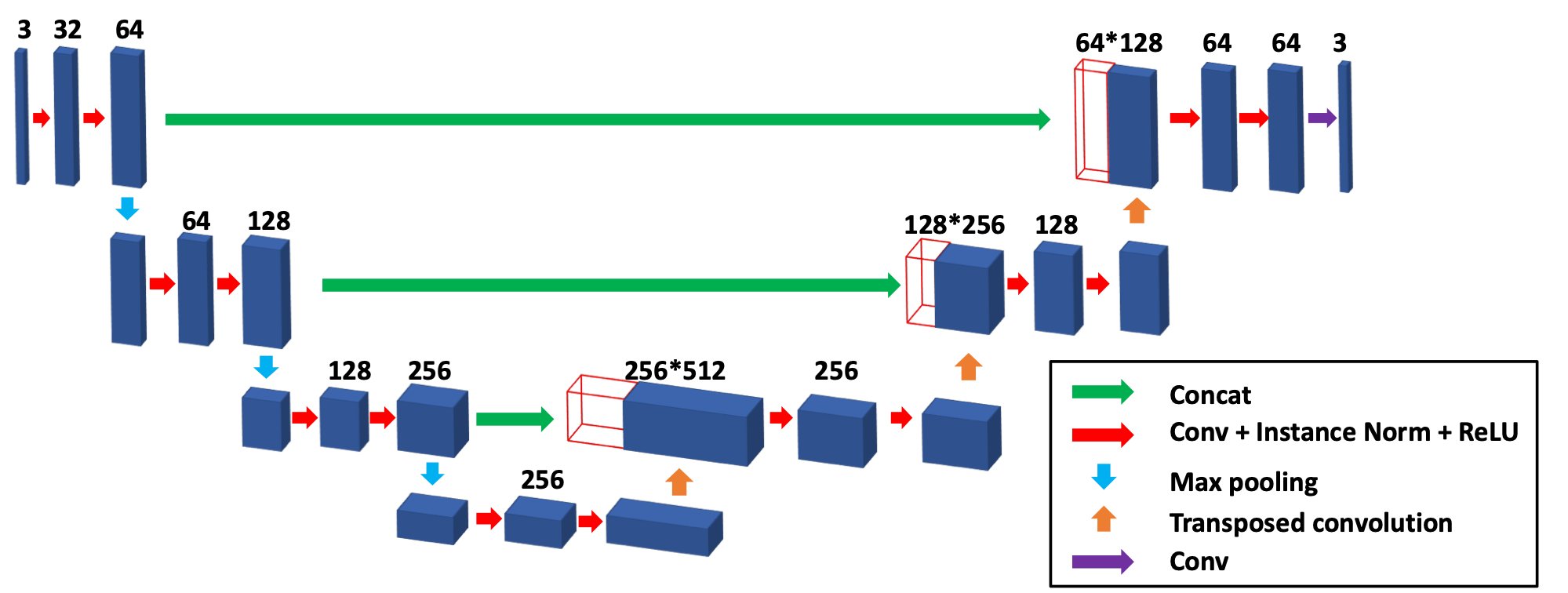}
\caption{3D U-net architecture for our generator.  
}
\label{fig:Unetwork}
\end{figure}

\begin{figure}[h!]
\centering\includegraphics[width=0.8\textwidth]{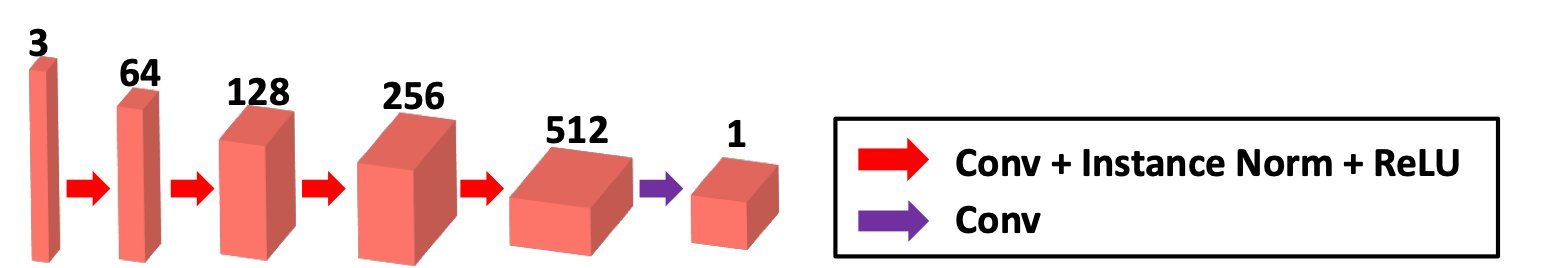}
\caption{3D discriminator architecture. The discriminator consists of 4 modules which consist of  {Conv} +  {Instance Norm} +  {ReLU}. Every  {Conv} layer has stride 2, and downsamples the input volume. At last layer, the number of output channel is 1.}
\label{fig:patchGANArchitec}
\end{figure}

\begin{figure}[h!]
\centering\includegraphics[width=0.7\textwidth]{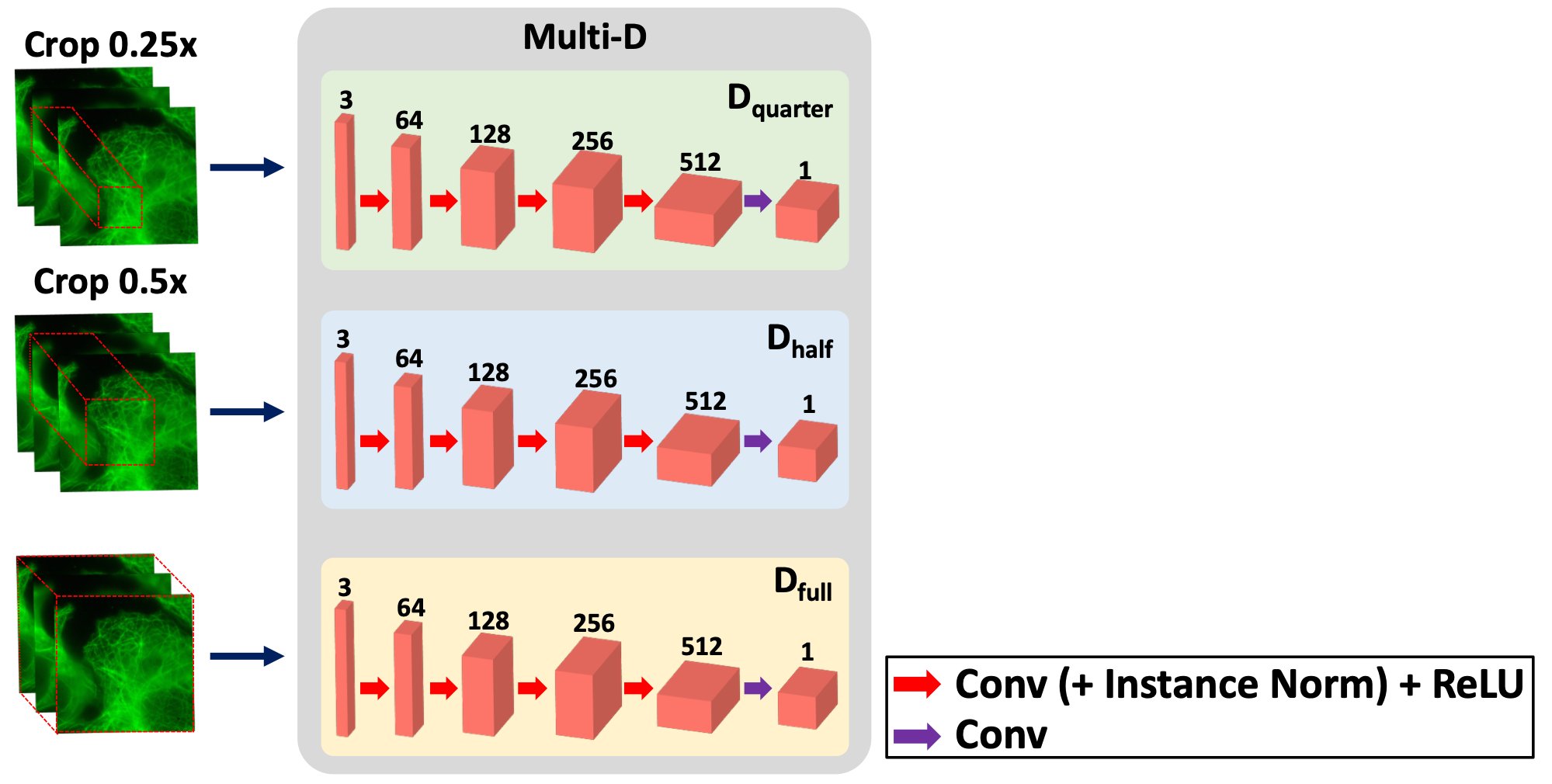}
\caption{Our multiple discriminators  consist of three independent discriminators. Each discriminator takes patches at different scales. Specifically, $D_{full}$ takes the patch in its original size; $D_{half}$ takes the randomly cropped patch half size of the original patch size; and $D_{quarter}$ takes the randomly cropped patch a quarter of the original patch size.}
\label{fig:multiPatchGANArchitec}
\end{figure}

\section{Network architecture}
The network architecture of the generator $G_{AB}$ is 3D-Unet \cite{cciccek20163d} as illustrated in Fig \ref{fig:Unetwork}.
For the architecture of $G_{AB}$, our U-net structure consists of contracting, expanding paths. The contracting path consists of the repetition of the following blocks: {3D conv}- {Instance Normalization} \cite{ulyanov2016instance}- {ReLU}.
Through the network, the convolutional kerenl dimmension is 3x3x3. At the first layer, a channel of the feature map is 64.
The network architecture of discriminators $D^{f_{i}}_{A},D^{f_{i}}_{B}$ is illustrated in Fig. \ref{fig:patchGANArchitec}. The discriminators are PatchGANs \cite{isola2017image}, and we use 3 discriminators that process patches with 3 different scales  as shown in Fig \ref{fig:multiPatchGANArchitec}. The network architecture of the discriminators consist of modules, which  consist of  {3D conv}- {Instance Normalization}- {ReLU}. Through the network, the convolution kernel dimmension is 3x3x3. 

On the other hand, the generator $G_{BA}$ uses only a single 3D  convolution layer to model a 3D blurring kernel. 
The size of the 3D PSF modeling layer is chosen depending on training set. 


\section{Method}
For training, we used 19 epifluorescence (EPF) samples of tubulin with a size of $512\times 512\times 30$. As for unmatched sharp image volume,
we use  deblurred image generated by utilizing a commercial software AutoQuant X3 (Media Cybernetics, Rockville). 
The volume depth is increased to 64 by padding with  {reflect}. Due to memory limitations, the volume is split into 64x64x64 patches. For data augmentation, rotation, flip, translation, and scale are imposed on the input patches. Adam optimizer with $\beta_{1}$=0.9 and $\beta_{2}$=0.999 is used to optimize the equation (\ref{eq:1}), and the learning rate is 0.0001. The learning rate decreases linearly after epoch 40; and the total number of epoch is 200. To reduce model oscillation \cite{goodfellow2014generative}, the discriminators used a history of generated volumes from a frame buffer containing 50 previously generated volumes. For all experiments, we set $\lambda_{1}$ of \eqref{eq:1} as 3 and $\lambda_{2}$ as 0.01. For the optimizer, we used only a single batch. We normalized the patches and set them to [0,1]. The PSF size is set to 20. The proposed method was implemented in Python with Tensorflow, and GeForce GTX 1080 Ti GPU was used for both training and testing the network.

To verify the performance of the proposed method, we compare  our method with commercial  deconvolution method using AutoQuant X3, supervised learning \cite{mao2016image},  and  the original cycleGAN  \cite{lu2017conditional} with both multi-PatchGANs and $G_{BA}$ from another CNN (Non-PSF layer).
In contrast to Lu et al. \cite{lu2017conditional} using regular CNN, our proposed model only used single PSF modeling layer in $G_{BA}$, making the training process much easier. 
 For supervised learning network, we  trained a 3D U-net with the matched label data from AutoQuant X3
 using  L1-loss since L1-loss encourages less blurring \cite{isola2017image}. 
All the reconstruction results were post-processed for better visualization by adaptive histogram equalization \cite{pizer1987adaptive}.

\begin{figure}[h!]
\centering
\begin{subfigure}[b]{0.9\textwidth}
\includegraphics[width=1\textwidth]{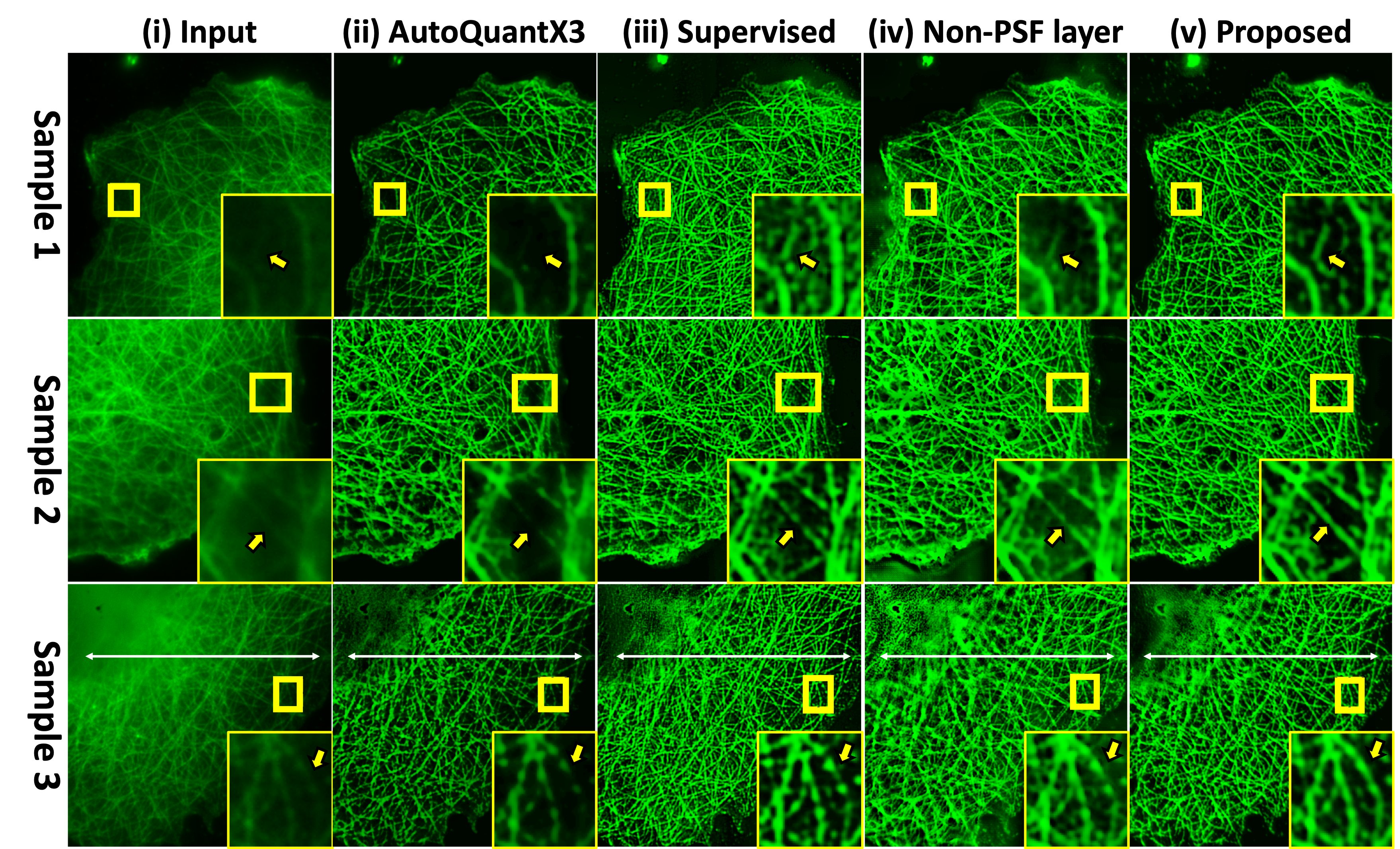}
\caption{Comparison of 3 test samples of transverse view over the follwing methods: AutoQuantX3, supervised learning, CycleGAN with both multi-PatchGANs and non-PSF layer (Non-PSF layer), and the proposed method. ROI (marked yellow) in lower right corner shows enlarged result.}
\hfill\hfill
\label{fig:topResult}
\end{subfigure}
\begin{subfigure}[b]{0.9\textwidth}
\includegraphics[width=1\textwidth]{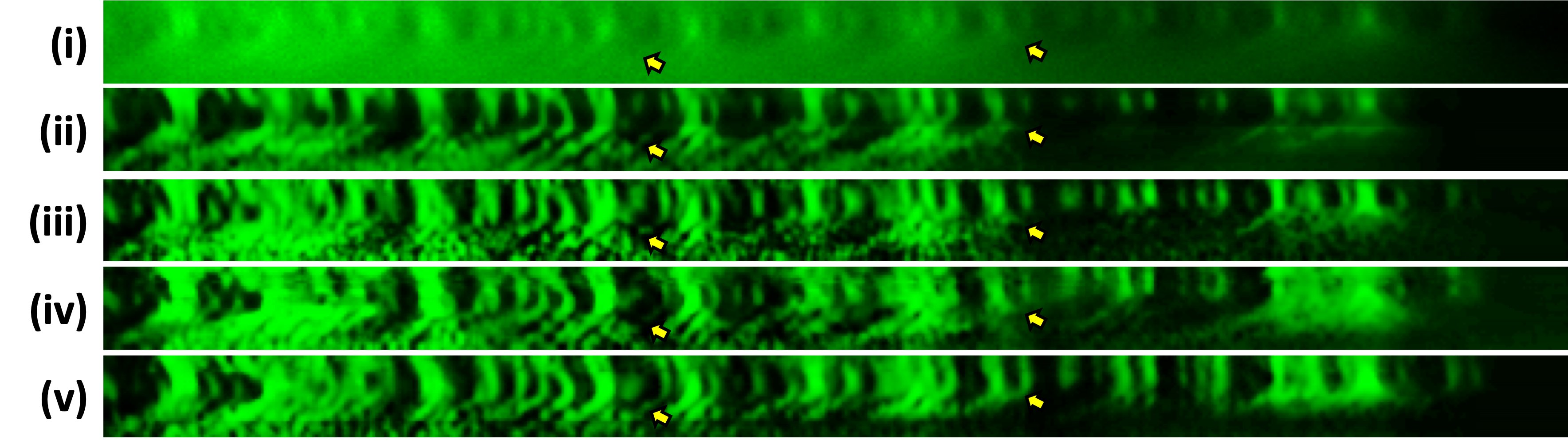}
\caption{Sagittal views of Sample 3 from Fig. \ref{fig:topResult}. The marked white line on sample 3 from Fig. \ref{fig:topResult} shows the scan line of the presented sagittal views.}

\label{fig:bottomResult}
\end{subfigure}
\caption{Result of transverse view and sagittal view.}
\label{fig:result}
\end{figure}
\section{Experimental Results}

{
Fig. \ref{fig:topResult}  and Fig. \ref{fig:bottomResult} show cross-views and sagittal views of  various reconstruction
method. 
In Fig. \ref{fig:topResult}, input images are degraded by blur and noise. Besides, as shown in Fig. \ref{fig:bottomResult}, the degradation at deeper depth gets worse. In Fig. \ref{fig:topResult}, AutoQuant X3 removed blur and noise; however, it did not improve contrast sufficiently. Both the supervised learning and the non-PSF layer showed better contrast and removed blur; however, the structural continuity was not preserved. In Fig. \ref{fig:bottomResult}, the AutoQuant X3, the supervised learning, and the non-PSF layer somehow removed blur and noise, but did not maintain structure continuity at deeper depth. Finally, in the proposed method blurs and noise were successfully removed in both Fig. \ref{fig:topResult} and Fig. \ref{fig:bottomResult}, thereby preserving the continuity of the structure. 
We therefore confirm that PSF modeling layer improves the robustness of the proposed method.   }

\section{Discussion and Conclusion}
In this paper, we presented a novel blind deconvolution using an unsupervised deep neural network using CycleGAN architecture. Experimental results showed that our proposed method restores the good quality reconstruction in both transverse and sagittal view. In particular, we observed that the use of PSF modeling layer improved the effectiveness of the proposed method. We have also proposed multiple patchGANs taking patches at different scales to discriminate real samples from generated results. The multiple patchGANs helped generators to produce coarsest and finest details. 
\end{document}